# Machine Unlearning using a Multi-GAN based Model

Amartya Hatua,[1, a)] Trung Nguyen,[1, b)] and Andrew H. Sung[3, c)]

[1] *Fidelity Investments Boston, MA 02210, USA*
[2] *Winona State University Winona, MN 55987, USA*
[3] *The University of Southern Mississippi Hattiesburg, MS 39401, USA*

[a)] *Corresponding author: amartyahatua@gmail.com*
[b)] *trung.nguyen@winona.edu*
[c)] *andrew.sung@usm.edu*

**Abstract.** This article presents a new machine unlearning approach that utilizes multiple Generative Adversarial Network (GAN) based models. The proposed method comprises two phases: i) data reorganization in which synthetic data using the GAN model is introduced with inverted class labels of the forget datasets, and ii) fine-tuning the pre-trained model. The GAN models consist of two pairs of generators and discriminators. The generator discriminator pairs generate synthetic data for the retain and forget datasets. Then, a pre-trained model is utilized to get the class labels of the synthetic datasets. The class labels of synthetic and original forget datasets are inverted. Finally, all combined datasets are used to fine-tune the pre-trained model to get the unlearned model. We have performed the experiments on the CIFAR-10 dataset and tested the unlearned models using Membership Inference Attacks (MIA). The inverted class labels procedure and synthetically generated data help to acquire valuable information that enables the model to outperform state-of-the-art models and other standard unlearning classifiers.

## INTRODUCTION

Machine learning is being used extensively in various applications that affect our daily lives. As a result, there is a growing concern about ethics and legality. Governments worldwide are taking steps to regulate the use of AI and ensure that algorithmic decisions and data usage practices are safe and ethical [1]. The World Economic Forum (WEF) [2] and the Canadian Directive in Automated Decision-Making (CDADM) [3] have both published procurement checklists for AI-based systems. The WEF guidebook offers practical advice for the public sector and has been evaluated in several countries, including the United Kingdom, Bahrain, the United Arab Emirates, India, and Brazil [4]. CDADM is one of the earliest regulations specifically targeting AI systems, and it became effective in April 2019. It mandates full compliance by October 2023 for all new systems and by April 2024 for all existing systems.

Many privacy regulations, including the European Union's General Data Protection Regulation (GDPR) [5], the California Consumer Privacy Act (CCPA) [6], and Canada's proposed Consumer Privacy Protection Act (CPPA) [7], require the deletion of private information [8]. To meet these requirements, specific data samples must be removed from the training dataset, or their effects on the trained model must be removed. Xu et al. [8] conducted a recent survey on machine unlearning, which revealed that many concerned parties had requested this type of data removal. Additionally, machine unlearning approaches can help improve model performance by removing outlier training samples.

The motivation of this paper is the first Machine Unlearning Challenge of Google running from July to September 2023 [9]. The main objective of this challenge is to find standard metrics to evaluate unlearning algorithms and discover new challenges or opportunities in this research domain. In this challenge, we are given a pre-trained model resulting from training a machine learning model on an original dataset. We need to develop an unlearning algorithm to modify such a pre-trained model to remove the effect of the samples in the forget dataset while maintaining its performance on the retain dataset. The retain dataset is the original dataset after removing the samples in the forget dataset. The success of an unlearning algorithm is tested based on the success of avoiding the inference of forget samples using MIA techniques and the similar model distribution of the fine-tuning model vs. the model to be trained from scratch on the retain dataset.

In this paper, we proposed a new machine unlearning approach to remove the effect of specific training samples, called forget dataset, on an existing pre-trained model. The fine-tuning model after retraining with our approach can approximate the performance of the model if we train only on the retain dataset. In summary, below are our contributions:

1. A new machine unlearning approach based on inverted class labels on forget dataset.
2. A study of applying Generative Adversarial Network (GAN) [10] techniques in improving the unlearning process by generating new samples for the forget and retain datasets.
3. An analysis of Membership Inference Attacks (MIA) [11] techniques to verify the performance of fine-tuning models after machine unlearning processes.

## RELATED WORKS

According to Xu et al., [8], the machine unlearning problem can be formed as follows. Let $D$ be the cluster of training samples to be trained with a learning algorithm $A$ to produce a model $A(D)$. Let's call $F$ the set of samples to be removed from $D$ and $A(D)$, a.k.a. forget dataset. Then, the unlearning algorithm $U(A(D), F, D - F)$ is a training process to transform the model $A(D)$ to forget the effects of samples in $F$ while maintaining its performance on retain set, $D - F$. The unlearned model resulting from the above process should behave similarly to the model trained on the dataset without the forget set from scratch. According to the authors, there are two general approaches to unlearning: data reorganization and model manipulation.

In the approach of data reorganization, the original training dataset undergoes modification through obfuscation, pruning, or replacement, as described in [8], before fine-tuning the existing model to obtain the unlearned model. In the obfuscation method, the forget samples are changed in a way that fine-tuning the pre-trained model with additional iterations can help the model "forget" such samples. In the pruning method, the original training dataset is split into multiple disjoint sub-datasets. Submodels are trained on each sub-dataset, and the final model is a combination of submodels using ensemble learning or voting. To unlearn forget samples, the submodels must be retrained on the affected sub-datasets after removing any instances of the forget samples. In the replacement method, the original dataset is transformed into a new dataset, making it easier to implement the unlearning algorithm. However, these three methods can break the correlation between samples and models and do not guarantee the unlearned model's performance on the retained dataset.

Compared to data reorganization, model manipulation methods are a more popular technique for unlearning approaches. The main idea of these methods is to estimate the influence of forget samples and remove such influence in the pre-trained model to get the unlearned model. There is research to be done on simple machine learning models, such as linear, logistic, or complex ones with special assumptions [12, 13, 14]. There is other research attempting to update the gradient [15] or parameter shifting [16] in unlearning training processes. To the best of our knowledge, for very complex models such as deep neural networks, there are no good ways to estimate the influence of specific samples in multiple weights and parameters of different layers of the models.

## PROPOSED METHODOLOGY

We introduced a new machine unlearning method that utilizes a multi-GAN approach inspired by two GAN-based Positive Unlabeled (PU) learning models proposed by Hou et al. [17] (GenPU) and Yang et al. [18]. Figure 1 shows the conceptual idea of our proposed model. Research [19, 20] has shown that in cases where data is lacking, synthetic data can be used to enhance the performance of classifiers. This has led to the idea of using synthetic data to address the problem of machine unlearning. In this scenario, a model is initially trained using a training dataset. The training dataset is then divided into forget and retain datasets, which are both applied to the same trained model to fine-tune forget and retain. However, the retain and forget datasets are smaller than the training dataset, which creates a data deficiency problem. To resolve this issue, synthetic data is added to the original data. The proposed method starts with a pre-trained machine-learning model $M$, forget dataset $F$, and retain dataset $R$. The objective of our approach is to modify such pre-trained model $M$ to unlearn certain data in $F$ that it has previously learned while its performance on the retain dataset $R$ should be preserved. Our proposed model consists of two main phases below:

### Data reorganization with GAN and inverted forget labels

The forget and retain datasets are employed to train two separate GAN models, $GAN_f$, $GAN_r$, which are capable of generating synthetic forget and retain datasets $D_{sf}$ and $D_{sr}$. Sample of real and synthetically generated images for retain dataset and forget dataset are presented in Fig. 2.

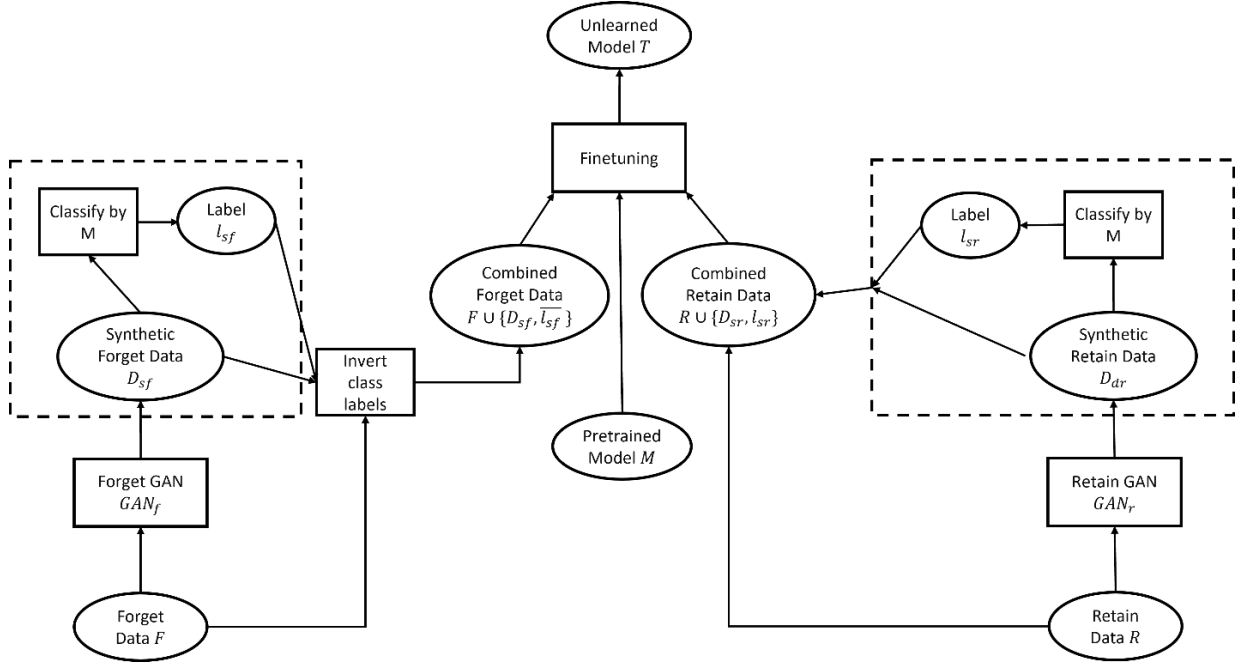

**FIGURE 1.** Schematic diagram of proposed model

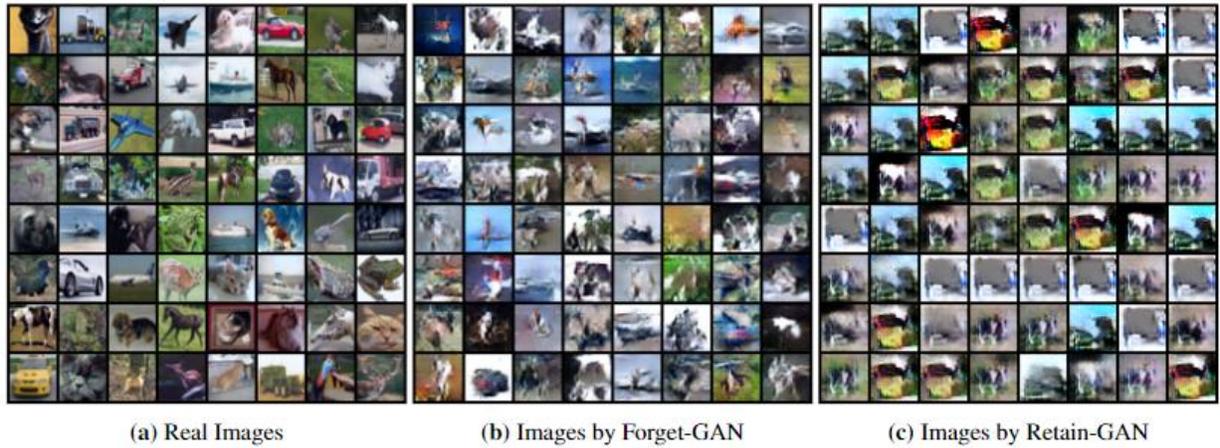

**FIGURE 2.** Sample of real and synthetically generated images

These synthetically generated datasets are added to the original retain and forget datasets during the unlearning process in which finetuning is performed on the pre-trained model $M$. The synthetic data generated by the GANs does not have the class labels. So, class labels are needed to use this data for the unlearning phase. Hence, the pre-trained model is used as a classifier, and the labels of both synthetic datasets $l_{sf}$ and $l_{sr}$ are obtained. After that, we will modify the class labels of the original and synthetic forget datasets (which we call the inverting process). There are multiple options for this inverting process: randomly select a different class label compared to the current ground truth one or take the complement class label, etc. There are two motivations behind this process. Firstly, using different class labels of forget samples requires the model weights to be changed accordingly in the backpropagation learning process in finetuning, which, in the end, removes the influence of the forget samples in our pre-trained model. Secondly, using these new class labels during the fine-tuning phase later can help our target unlearned model avoid being attacked by MIA models. Steps 1 and 2 in our proposed Algorithm 1 explain how to train the GAN models to generate the synthetic data. We develop two GAN models: $GAN_f$ generates synthetic forget data and $GAN_r$ generates synthetic retain data. $G_f$ is responsible for generating forget data, and $D_f$ discriminates between original forget and synthetically generated forget data. $G_r$ and $D_r$ are responsible for similar functions for retain data. While step 3 in our proposed Algorithm 1 explains how to generate the class labels for both synthetic forget and retain datasets, step 4 describes the process of obtaining the inverted labels for the forget datasets and the ground truth class labels of the retain datasets.

### Fine-tuning the pre-trained model with combined datasets

After the previous phase, we can develop two combined datasets: original/synthetic forget datasets with inverted labels and original/synthetic retain datasets with correct labels. We perform fine-tuning the pre-trained model using additional epochs on the two combined datasets above. In each epoch, we start by training the model to "forget" the samples in the combined forget dataset, and then optimize its performance loss on the combined retain dataset. Our argument is that while updating the model weights to produce different class labels compared to the previous ones they learned before, the performance will drop. Therefore, the second optimization step on the retain datasets is required to recover its performance. This phase is presented in step 5 of our proposed Algorithm 1.

---

**Algorithm 1** Unlearning through fine-tuning models with inverted forget set and GAN data

**Inputs**: Forget dataset $F$, Retain dataset $R$, A pre-trained model $M$
**Step 1**: Training the forget GAN model $GAN_f$
1: **for** each sample in the forget dataset $F$ **do**
2:     update forget discriminator network $D_f$
3:     update forget generator network $G_f$
4:     $min_{G_f} max_{D_f} V(D,G) = min_{G_f} max_{D_f} E_{x \sim p_f(x)} \log\left(D_f(x)\right) + E_{z \sim p_f(x)} \log\left(1 - D_f\left(G_f(z)\right)\right)$
5: **end for**
**Step 2:** Training the retain GAN model $GAN_r$
6: **for** each sample in the retain dataset $R$ **do**
7:     update retrain discriminator networks $D_r$
8:     update retain generator networks $G_r$
9:     $min_{G_r} max_{D_r} V(D,G) = min_{G_r} max_{D_r} E_{x \sim p_r(x)} \log(D_r(x)) + E_{z \sim p_r(x)} \log\left(1 - D_r(G_r(z))\right)$
10: **end for**
**Step 3**: Generate synthetic forget data $D_{sf}$ and synthetic retain data $D_{sr}$
11:   $D_{sf} = E_{z \sim p_f(x)} G_f(z)$
12:   $D_{sr} = E_{z \sim p_r(x)} G_r(z)$
**Step 4**: Classify synthetic data and invert the class labels of forget set
13:   $l_f = Invert\left(M(d_{sf})\right)$     ▷ inverted forget class labels
14:   $l_r = M(d_{sr})$     ▷ retain class labels
**Step 5**: Finetuning the pre-trained model $M$ with two combined datasets
15: Target model $T = M$
16: **for** each epoch in #epochs **do**
17:     Finetune $T$ using $F \cup \{D_{sf}, l_f\}$
18:     Finetune $T$ using $F \cup \{D_{rf}, l_r\}$
19: **end for**
20: **Return the unlearned model $T$**

---

## EXPERIMENTAL RESULTS AND DISCUSSIONS

As per the guidelines of the NeurIPS 2023 Machine Unlearning Challenge [21], we conducted experiments using the CIFAR-10 [22] dataset, which is divided into training, testing, forget, and retain datasets. The training set contains 50,000 images, whereas the test set consists of 5,000 images. Afterward, the training set is further divided into the forget set and retain set, which contain 5000 and 45000 images, respectively. For the GAN models of the forget and retain datasets, we used similar generator and discriminator models. During the training process, we compared the distribution of synthetic and real data using KL divergence after every epoch. In Table 2, KL divergence of different epochs is presented. We can observe that the KL divergence for both the GAN models decreases over epochs. Fig 3 shows the KL divergence values for different epochs for both models. Once the GANs are trained, the generators can produce synthetic data with a similar distribution to the original data when we input random data into them. In the fine-tuning phase of the target model, we used these generators to generate synthetic data. This synthetic data is eventually used to finetune the target model for unlearning. We compared the results of three models: Baseline 1, which was fine-tuned using only the Retain dataset [21], Baseline 2, which was fine-tuned using the Retain and Forget datasets processed with our approach without the synthetic GAN components, and our proposed model after fine-tuning the pre-trained model on the combined original/synthetic datasets after processing using our approach. The performance comparison of the three models is presented in Table 1.

**Tabel 1:** Results of the proposed models and two baseline models

|                | Retain accuracy | Test accuracy |
|----------------|-----------------|---------------|
| Base line 1    | 98.1%           | 83.2%         |
| Base line 2    | 97.5%           | 86.1%         |
| Proposed Model | 97.7%           | 86.1%         |

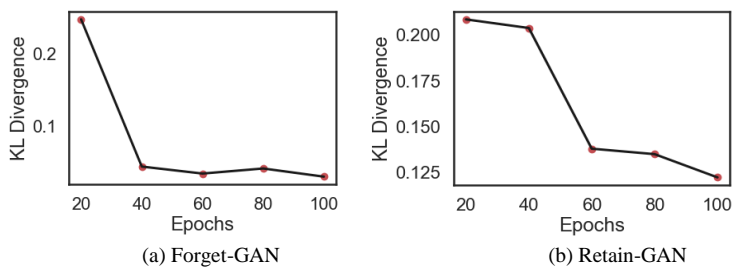

(a) Forget-GAN      (b) Retain-GAN

**Figure 3:** KL Divergence for Forget-GAN and Retain-GAN for different epochs

Before the fine-tuning phase of the target model in our experiments, we utilized generators to create synthetic data. To determine the class labels for the synthetic data, we employed a pre-trained ResNet-18 [23] classifier (this is the same pre-trained classifier where the unlearn algorithm is actually applied). For the retain set, the class label is unchanged, while for the forget synthetic data, we complemented 9 to the original class labels to obtain new ones. Both the synthetic datasets were combined with the original datasets to create a larger dataset for fine-tuning the target model in the second phase. We repeated this synthetic data generation process multiple times, gradually increasing the volume of the data. Finally, we performed the fine-tuning of the pre-trained ResNet-18 model using the combined datasets with increasing numbers of epochs.

**Tabel 2:** Results of the proposed models and two baseline models

|               | Epochs |       |       |       |       |
|---------------|--------|-------|-------|-------|-------|
| KL Divergence | 20     | 40    | 60    | 80    | 100   |
| Forget GAN    | 0.208  | 0.203 | 0.137 | 0.134 | 0.122 |
| Retain GAN    | 0.248  | 0.042 | 0.033 | 0.040 | 0.028 |

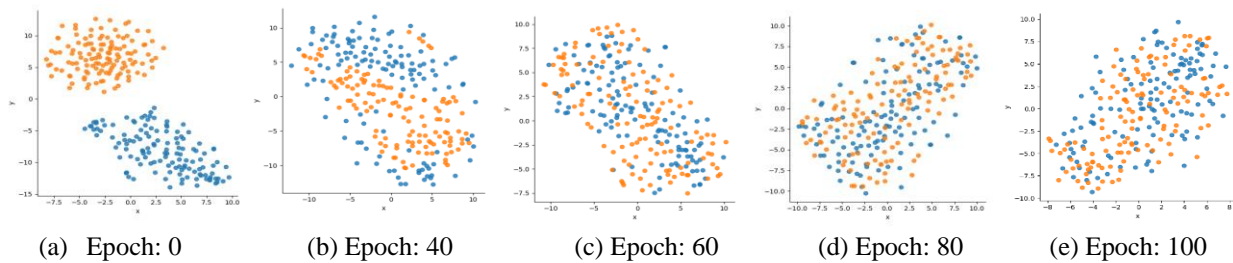

(a) Epoch: 0    (b) Epoch: 40    (c) Epoch: 60    (d) Epoch: 80    (e) Epoch: 100

**Figure 4:** t-SNE plot of Real and Synthetic Forget Data (Orange and Blue colored points are representing Real and Synthetic data respectively)

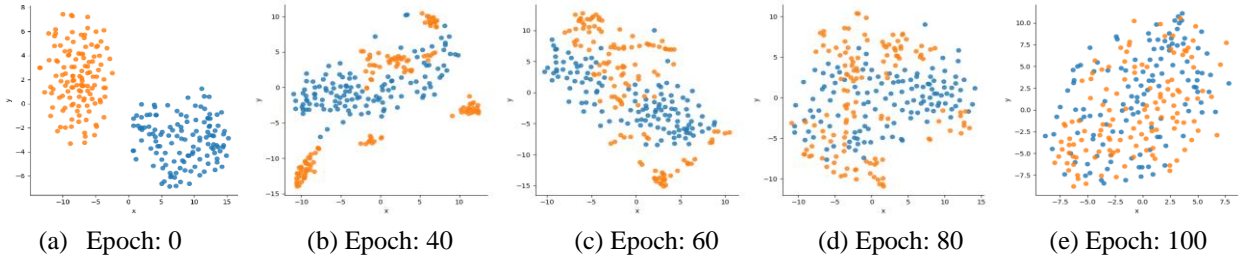

| (a) Epoch: 0 | (b) Epoch: 40 | (c) Epoch: 60 | (d) Epoch: 80 | (e) Epoch: 100 |

**Figure 5:** t-SNE plot of Real and Synthetic Retain Data (Orange and Blue colored points are representing Real and Synthetic data respectively)

In addition to measuring accuracy, we also utilized MIA as a metric to evaluate the "unlearning" performance and privacy preservation capability of our model. MIA assesses whether the loss distribution of the model's training and test sets are comparable. To conduct this test, we employed three different classifier models, namely Linear Logistic Regression (LR) [24], Support Vector Machine (SVM) [25], and XGBoost (XGB) [26]. These models were used to classify and compare the loss distributions. We have observed the MIA score for all the classifiers using the pre-trained ResNet-18 model, Base model 1, Base model 2, and our proposed model. It is presented in Table 3. Based on the MIA scores of a pre-trained model, it appears that classifiers can easily differentiate the loss distribution between the test set and the forget dataset. This implies that the pre-trained model is susceptible to MIA. On the other hand, the scores for unlearned models are closer to 0.50, indicating that classifiers are unable to discern the difference in loss distribution between the test and forget dataset. Consequently, unlearning effects are applied, making the unlearned models less susceptible to MIA.

**TABLE 3.** MIA Score for different models

|  | Logistic Regression | SVM | XGBoost |
|---|---|---|---|
| Pretrained ResNet-18 | 0.577 | 0.576 | 0.577 |
| Base line 1 | 0.510 | 0.507 | 0.503 |
| Base line 2 | 0.533 | 0.533 | 0.531 |
| Proposed model | 0.503 | 0.504 | 0.508 |

## LIMITATIONS

Our proposed unlearning methodology provides a practical approach to unlearn specific samples in the forget dataset from a pre-trained model while maintaining its performance on the retain dataset. We boost the unlearning process with more generalized synthetic data from the original forget and retain datasets. To the best of our knowledge, here are some possible limitations of this approach:
- The GAN models are known to be unstable for some specific datasets. In such a case, the boosting effect of synthetic data may not be achieved.
- Training the GAN models for the dataset may take a significant amount of time and slow down the unlearning process if we need to find the unlearned in a short time.
- In this study, we haven't tested our method on other kinds of datasets, so there may be performance issues or challenges that we haven't seen.

## FUTURE WORK

One promising direction for further research involves applying the algorithm to different datasets to evaluate its generalizability. Additionally, the use of Forgetting Neural Networks (FNNs), inspired by Mathematical Theorems Explaining Human Universal Forgetting, could facilitate machine unlearning. Cano et al. [27] proposed a mathematical model for the 'Theory of Forgetting' by implementing an FNN. Exploring whether human forgetting patterns can improve machine learning models would be a fascinating area to investigate. Moreover, leveraging the principles of differential privacy could lead to the development of machine unlearning algorithms.

# CONCLUSION

We propose a new unlearning approach for machine learning models based on data reorganization and model manipulation. In the first step, data reorganization, we perform data obfuscation on the forget dataset by modifying the class labels of its samples using the complement ones. Then, GAN models are used to train and generate synthetic data for forget and retain datasets. The class labels of samples in the synthetic forget set are obfuscated, similar to the original forget dataset. Finally, in the second phase of model manipulation, we perform fine-tuning the pre-trained model to reduce the effect of the samples on the original/synthetic forget datasets while optimizing the model on the performance of the original/synthetic retain sets. Because the class labels of both forget datasets are modified while fine-tuning the pre-trained model, their effects on the pre-trained model weights will be canceled/removed. However, doing so may lower the model's performance. Therefore, we optimize the model weights on the performance of retain samples.

Our experimental results on the CIFAR-10 dataset have shown the novelty of our approach in maintaining the model performance on the forget, retain, and test sets after fine-tuning. We also verified the "forget" capability of the fine-tuned models using Membership Inference Attack (MIA) models. We built three MIA models (LR, SVM, and XGBoost) to predict whether the fine-tuned models are trained with specific data samples based on the loss when testing such samples. Our MIA scores were in the range of 0.5 when testing on our unlearned model, which indicated the failure to infer the forget samples from our models.

Our approach has some limitations, e.g., the instability of GAN for specific datasets, time for training the GAN models, and the lack of completeness, as we have not tested on other kinds of data. However, our approach has shown promising results on image datasets trained with deep learning models and can be generalized to other types of data as well. In the future, we would like to explore the differential unlearning approach in which estimation of the influence of training samples on the model weights parameters is conducted, and difference approximation will take place to adjust the model weights to reflect the unlearned samples. Other than that, unlearning some specific features and their values can be explored rather than a set of forget samples. Besides, more GAN ideas can be exploited in the unlearning process. Lastly, different advanced MIA models can be used as optimized objectives when fine-tuning the target model.